\documentclass{article}

\usepackage{microtype}
\usepackage{graphicx}
\graphicspath{{figures/}}
\usepackage{subcaption}
\usepackage{booktabs}
\usepackage{hyperref}

\usepackage[preprint]{icml2026}

\usepackage{amsmath}
\usepackage{amssymb}
\usepackage{mathtools}
\usepackage{amsthm}
\usepackage{algorithm}
\usepackage{algorithmic}
\usepackage[table]{xcolor}
\usepackage{multirow}

\usepackage[capitalize,noabbrev]{cleveref}

\icmltitlerunning{Striding Across Reynolds Numbers}

\begin{document}

\twocolumn[
  \icmltitle{Striding Across Reynolds Numbers: \\ Representation Geometry in Neural PDE Generalisation}

  \icmlsetsymbol{equal}{*}

  \begin{icmlauthorlist}
    \icmlauthor{Jianing Shi}{lse}
  \end{icmlauthorlist}

  \icmlaffiliation{lse}{London School of Economics and Political Science}
  \icmlcorrespondingauthor{Jianing Shi}{J.Shi39@lse.ac.uk}

  \vskip 0.3in
]

\printAffiliationsAndNotice{}

\begin{abstract}
Cross-Reynolds generalisation in neural PDE solvers remains poorly characterised.
On the canonical forced 2D Navier--Stokes benchmark, a trained Fourier Neural Operator reaches $46.68\%$ relative $L_2$ error under a $10{\times}$ Reynolds-number shift, yet zero-forward-model retrieval baselines already improve to $41$--$42\%$.
This suggests representation geometry as a major organising variable among the tested methods.
We test this hypothesis through ConvAE-Relay, which matches states in a source-trained convolutional autoencoder latent space and borrows dynamics from a source-regime database, achieving $38.34{\pm}0.07\%$ using only a source-regime database and no target-regime fitting, labels, or database entries.
A $2{\times}2$ ablation isolates matching quality as dominant over the update rule.
Oracle experiments confirm that source-regime dynamics directions remain transferable (cosine similarity ${\sim}0.84$) when matching stays on-manifold; autoregressive drift is the primary bottleneck (${\sim}12$ percentage points).
From the learned-prediction side, a U-Net with multi-scale skip connections achieves $34.72{\pm}0.60\%$, consistent with the retrieval-side finding that local, multi-scale representations organise cross-Reynolds transfer among tested methods.
All claims are scoped to this benchmark.
\end{abstract}

\section{Introduction}\label{sec:intro}

A zero-forward-model analogue relay outperforms a trained Fourier Neural Operator under a $10{\times}$ Reynolds-number shift.
Yet a learned U-Net predictor outperforms both.
This apparent contradiction, retrieval beating spectral learning while a different learned architecture beats retrieval, points to a shared underlying variable: the geometry of the state representation used for cross-regime prediction.

The Reynolds number $\mathrm{Re} = UL/\nu$ governs the ratio of inertial to viscous forces; increasing Re activates finer spatial scales through weakened viscous damping \citep{pope2000turbulent}.
Extrapolation across Re regimes is ubiquitous in engineering: wind-tunnel data applied to full-scale aerodynamics, moderate-resolution climate simulations extrapolated to finer regimes.
This is not arbitrary covariate shift; it is extrapolation along an axis whose difficulty is governed by the Navier--Stokes equations themselves.
Machine learning for fluid mechanics has advanced rapidly \citep{brunton2020machine, duraisamy2019turbulence}, yet cross-parameter generalisation remains underexplored.

On the forced 2D Navier--Stokes benchmark of \citet{li2021fourier}, we evaluate six methods spanning learned predictors and zero-parameter retrieval.
Three FNO-family variants concentrate at $46$--$48\%$ relative $L_2$ error at $\mathrm{Re}{\approx}10{,}000$, suggesting a persistent limitation of tested FNO-family protocols under this shift, although we do not claim exhaustive FNO optimisation.
PCA relay ($41.77\%$) and $k$NN-copy ($41.01\%$), both zero-parameter methods, outperform FNO.
This raises the question: is the bottleneck in the learned forward dynamics map, or in how states are represented?

PCA relay shows that retrieval can transfer useful dynamics, but PCA imposes a global linear geometry on vorticity fields.
We test whether a source-trained convolutional autoencoder provides a more indexable nonlinear representation: ConvAE-Relay achieves $38.34{\pm}0.07\%$, an improvement of $3.4$ percentage points (pp) over PCA relay with only the matching representation changed.
From the learned-prediction side, a U-Net with skip connections achieves $34.72{\pm}0.60\%$, the strongest tested result, while contrasting with the original in-distribution comparison where U-Net underperformed FNO \citep{li2021fourier}.
Retrieval-side ablations and learned-predictor comparisons provide consistent evidence from two complementary perspectives that representation geometry organises cross-Re transfer.

A $2{\times}2$ ablation (matching space $\times$ update rule) isolates matching quality as the dominant factor.
Oracle experiments decompose the error into drift (${\sim}12$\,pp), magnitude (${\sim}3$\,pp), and a diagnostic floor ($23.32\%$).
Mechanism ablations (smart tracking, dynamics-aware AE, history matching) each fail to improve, sharpening the diagnosis by elimination.
Boundary characterisation at $\mathrm{Re}{=}100{,}000$ reveals that relay's advantage reverses, while long-horizon evaluation shows drift accumulation eroding matching quality by $T{=}20$.

Our contributions are:
\textbf{(1)}~On this benchmark under $10{\times}$ Re shift, tested methods organise by representation geometry, with local and multi-scale representations outperforming global spectral or global linear.
\textbf{(2)}~ConvAE-Relay instantiates learned-representation analogue forecasting without a learned forward dynamics map, enabling controlled matching/update diagnostics.
\textbf{(3)}~Oracle and ablation experiments identify autoregressive drift as the main deployable bottleneck, expose a matching-dynamics representation tradeoff, and separate predictive sufficiency from relay indexability.

\section{Related Work}\label{sec:related}

\textbf{Neural operators.}
FNO \citep{li2021fourier} established the benchmark for learned PDE surrogates via spectral-domain convolutions.
Factorised variants \citep{tran2023factorized} and geometry-aware extensions \citep{li2023fourier} reduce cost or handle irregular domains.
The neural operator framework has been formalised as learning maps between function spaces \citep{kovachki2023neural}.
Broader families, including DeepONet \citep{lu2021learning}, graph-based simulators \citep{sanchez2020learning, pfaff2021learning}, Transformer-based operators \citep{cao2021choose, hao2023gnot}, and multiwavelet operators \citep{gupta2021multiwavelet}, deliver strong in-distribution performance, but cross-parameter generalisation is less studied.

\textbf{Out-of-distribution generalisation in scientific ML.}
Physics-informed approaches \citep{raissi2019physics, karniadakis2021physics} embed PDE structure into training but primarily target interpolation.
Scale-consistent learning \citep{li2025scaleconsistent} modifies training through scale-aware augmentation.
\citet{subramanian2024towards} characterise scaling and transfer behaviour for scientific foundation models.
Our work instead studies zero-target-regime diagnostics across representation families, including nonparametric transition reuse without a learned forward map.
U-shaped neural operators (U-NO; \citealp{rahman2023uno}) exist but evaluate in different settings.
To our knowledge, no prior work reports systematic cross-Re zero-shot evaluation comparing spectral, retrieval, and multi-scale methods on the original forced FNO benchmark under a unified protocol.

\textbf{Autoregressive rollout strategies.}
Long-horizon neural PDE prediction suffers from error accumulation.
Pushforward training \citep{brandstetter2022message} and diffusion-based refinement \citep{lippe2024pde} address this from the training side.
Data-driven weather forecasting \citep{lam2023learning, pathak2022fourcastnet} demonstrates that autoregressive neural surrogates can reach operational accuracy.
Our oracle decomposition isolates drift as the dominant bottleneck, providing a diagnostic complement to these training-side solutions.

\textbf{Analogue forecasting.}
\citet{lorenz1969atmospheric} introduced prediction from historical atmospheric archives via state matching; modern variants apply analogue methods to precipitation nowcasting \citep{atencia2015analogs} and other geophysical fields.
$k$NN-LM \citep{khandelwal2020generalization} demonstrated that retrieval-augmented prediction can match or surpass trained language models, especially under domain shift.
Representation quality for retrieval has been studied extensively in NLP \citep{bengio2013representation} but not in PDE dynamics transfer.
ConvAE-Relay bridges this gap: learned-representation analogue forecasting adapted to cross-regime PDE dynamics.

\section{Problem Setup and Methods}\label{sec:methods}

\begin{figure*}[t]
\centering
\includegraphics[width=\textwidth]{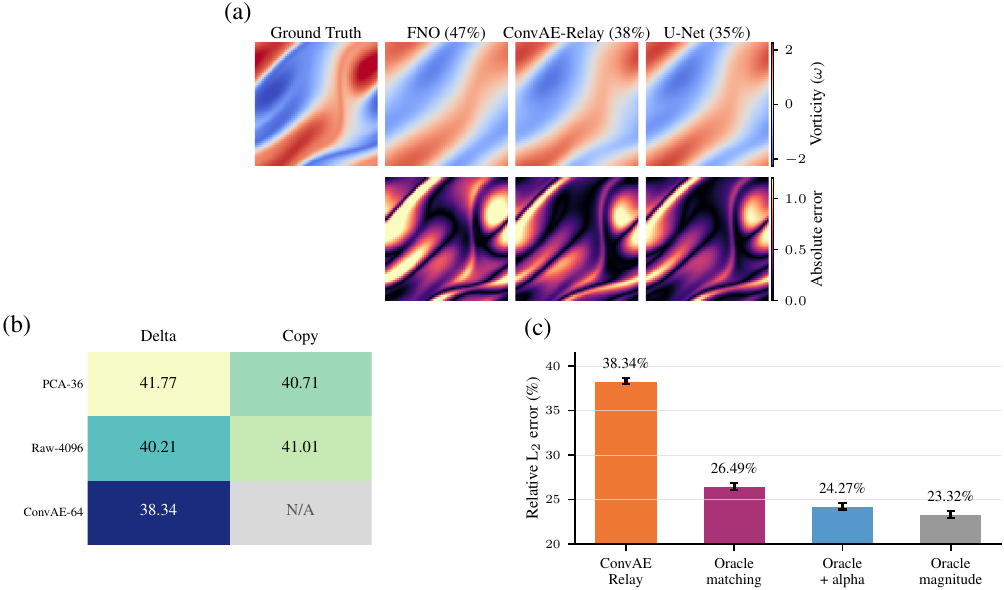}
\caption{Cross-Reynolds diagnostic summary.
\textbf{(a)}~Vorticity predictions and absolute error maps at $\mathrm{Re}{=}10{,}000$ (rollout step 5, median-error trajectory).
FNO produces visibly smoother fields; ConvAE-Relay and U-Net preserve finer structure.
\textbf{(b)}~$2{\times}2$ representation ablation (seed 42): upgrading matching space from PCA to ConvAE latent improves the best result by $2.37$\,pp, while the update-rule effect is ${\leq}1.1$\,pp; matching quality dominates.
\textbf{(c)}~Oracle bottleneck decomposition: autoregressive drift accounts for ${\sim}12$\,pp, magnitude mismatch ${\sim}2$\,pp, and residual ${\sim}1$\,pp.
Oracle variants are target-informed diagnostic probes, not deployable methods.}
\label{fig:hero}
\end{figure*}

\subsection{Benchmark and Task}\label{sec:benchmark}

We study the forced 2D vorticity Navier--Stokes equations on a $64{\times}64$ periodic domain \citep{li2021fourier,kraichnan1967inertial}:
\begin{equation}\label{eq:ns}
  \partial_t \omega + u\cdot\nabla\omega = \nu \Delta \omega + f, \qquad \nabla\cdot u = 0,
\end{equation}
where $\omega$ is vorticity, $u = \nabla^\perp \psi$ is velocity recovered from the stream function $\Delta\psi = -\omega$, $\nu$ is viscosity, and $f$ is Kolmogorov forcing at wavenumber $k_f{=}4$.
The benchmark fixes the characteristic velocity $U$ and domain length $L$, so regimes are indexed by inverse viscosity: $\mathrm{Re} = UL/\nu \propto \nu^{-1}$.
Table~\ref{tab:dataset} summarises the three regimes.
Lower viscosity reduces dissipation at small spatial scales, making cross-Re generalisation a test of whether learned representations preserve coarse-to-fine dynamical structure.

\begin{table}[t]
\caption{Dataset summary. All trajectories share the $64{\times}64$ periodic domain. The source split contains 5{,}000 available trajectories; learned predictors and relay databases use trajectories 0--3{,}999 for training, with trajectories 4{,}000--4{,}999 reserved for source-regime validation.}
\label{tab:dataset}
\vskip 0.1in
\centering
\small
\begin{tabular}{@{}llccl@{}}
\toprule
Regime & $\nu$ & Re & Traj. & Purpose \\
\midrule
Source & $10^{-3}$ & ${\approx}1{,}000$ & 5{,}000 & Train/val \\
Target 1 & $10^{-4}$ & ${\approx}10{,}000$ & 500 & OOD eval \\
Target 2 & $10^{-5}$ & ${\approx}100{,}000$ & 500 & Boundary \\
\bottomrule
\end{tabular}
\vskip -0.1in
\end{table}

\textbf{Task.}
Given frames $0$--$9$ as input context, predict frames $10$--$19$ via 10-step autoregressive rollout: $\hat{\omega}_{t+1} = f_\theta(\hat{\omega}_{t-9:t})$, where each prediction feeds back as input to the next step, compounding any single-step error.
All learned predictors (FNO and U-Net) receive the same 10-frame stacked input; relay methods use only the most recent encoded state for matching but initialise from the same context window.
The evaluation metric is mean relative $L_2$ error:
\begin{equation}\label{eq:metric}
  \mathrm{RelErr} = \frac{100}{NT}\sum_{n=1}^{N}\sum_{t=1}^{T}
  \frac{\lVert \hat{\omega}^{(n)}_t - \omega^{(n)}_t \rVert_2}{\lVert \omega^{(n)}_t \rVert_2}\;\%,
\end{equation}
where $N{=}500$ trajectories and $T{=}10$ rollout steps.

\textbf{Data protocol.}
Training data: $\mathrm{Re}{\approx}1{,}000$, trajectories $0$--$3{,}999$, frames $0$--$49$.
Relay database: $\mathrm{Re}{\approx}1{,}000$, trajectories $0$--$3{,}999$, frames $10$--$49$.
Evaluation: $\mathrm{Re}{\approx}10{,}000$, trajectories $0$--$499$, frames $10$--$19$.
$\mathrm{Re}{\approx}10{,}000$ data is used at evaluation only; no target-regime fitting, labels, or database construction at the target regime.

\subsection{Database Relay}\label{sec:relay}

Given an encoder $\phi$ mapping vorticity fields to a representation space, let $z_t = \phi(\omega_t)$ denote the encoded state at step~$t$.
A relay database $\mathcal{D}_s = \{(\phi(\omega^s_j), \Delta\omega^s_j)\}$ stores source-regime states and their transitions $\Delta\omega^s_j = \omega^s_{j+1} - \omega^s_j$.
At inference, the relay retrieves the nearest source state and borrows its transition:
\begin{equation}\label{eq:relay-match}
  j^\star_t = \operatorname*{arg\,min}_{j \in \mathcal{D}_s} \; d\!\left(\phi(\omega_t),\, \phi(\omega^s_j)\right), \qquad
  \hat{\omega}_{t+1} = \omega_t + \Delta\omega^s_{j^\star_t},
\end{equation}
where $d(\cdot,\cdot)$ is cosine distance in representation space:
\begin{equation}\label{eq:cosine-dist}
  d_\phi(\omega_i, \omega_j) = 1 - \frac{\langle \phi(\omega_i),\, \phi(\omega_j) \rangle}{\lVert \phi(\omega_i) \rVert_2 \; \lVert \phi(\omega_j) \rVert_2}.
\end{equation}
The method succeeds only if the nearest neighbour in representation space has a transition vector useful at the target regime.
The process repeats autoregressively for $T{=}10$ steps; a ride length of $N_\mathrm{ride}{=}3$ reuses the same source trajectory for 3 consecutive steps before re-matching.

\textbf{PCA relay} sets $\phi$ to PCA with 36 components (retaining $99.77\%$ of $\mathrm{Re}{=}1{,}000$ variance).
This is a zero-learned-dynamics baseline.

\textbf{$k$NN-copy} matches in raw $4{,}096$-dimensional vorticity space and copies the matched source's next frame without delta computation, equivalent to the raw-field/raw-copy cell of the $2{\times}2$ ablation (Table~\ref{tab:2x2}).

\subsection{ConvAE-Relay}\label{sec:convae}

PCA relay demonstrates that source-regime analogues can transfer useful dynamics, but PCA imposes a global linear geometry on vorticity fields.
We test whether a source-trained convolutional autoencoder provides a more indexable nonlinear representation for nearest-neighbour dynamics borrowing.

The convolutional autoencoder consists of 4 stride-2 convolutional layers ($1{\to}32{\to}64{\to}128{\to}256$) with batch normalisation and ReLU, followed by a linear projection to a 64-dimensional latent space; a mirror decoder reconstructs the vorticity field.
It is trained on $\mathrm{Re}{=}1{,}000$ frames with MSE reconstruction loss for 50 epochs, achieving ${\sim}1\%$ reconstruction error on source data.

The database stores $156{,}000$ encoded source-regime transitions.
At evaluation, $\mathrm{Re}{=}10{,}000$ frames are encoded and matched in this latent space following Equation~\eqref{eq:relay-match}.
ConvAE-Relay does not learn a forward dynamics map; it is database-indexed transition reuse in a learned local representation.
The autoencoder has ${\sim}1.6$M parameters trained on source data; only the retrieval mechanism is nonparametric.
Algorithm~\ref{alg:relay} summarises the procedure.

\begin{algorithm}[t]
\caption{ConvAE-Relay inference}
\label{alg:relay}
\begin{algorithmic}[1]
\REQUIRE Target context $\omega_0,\ldots,\omega_9$; source database $\mathcal{D}_s$; encoder $\phi$; ride length $N_\mathrm{ride}$
\STATE $\omega \leftarrow \omega_9$ \COMMENT{initialise from last context frame}
\FOR{$t = 1$ \textbf{to} $T$}
  \IF{$t \bmod N_\mathrm{ride} = 1$}
    \STATE $j^\star \leftarrow \operatorname*{arg\,min}_{j \in \mathcal{D}_s}\; d(\phi(\omega),\, \phi(\omega^s_j))$ \COMMENT{re-match}
  \ELSE
    \STATE $j^\star \leftarrow j^\star + 1$ \COMMENT{ride source trajectory}
  \ENDIF
  \STATE $\hat{\omega} \leftarrow \omega + \Delta\omega^s_{j^\star}$ \COMMENT{borrow transition}
  \STATE Store $\hat{\omega}$ as prediction for step $t$
  \STATE $\omega \leftarrow \hat{\omega}$ \COMMENT{autoregressive feedback}
\ENDFOR
\end{algorithmic}
\textit{Decoder $\phi^{-1}$ used only during autoencoder training; inference uses the encoder for retrieval and applies raw vorticity-space source deltas. No target-regime fitting, labels, or database.}
\end{algorithm}

\subsection{U-Net Predictor}\label{sec:unet}

We train a U-Net \citep{ronneberger2015unet} with an encoder-decoder architecture and skip connections.
Circular padding respects the periodic domain.
Input: 10 frames stacked in the channel dimension; output: the next frame, $\hat{\omega}_{t+1} = \mathcal{U}_\theta(\omega_{t-9:t})$.
Base channels progress as $32{\to}64{\to}128{\to}256{\to}512$ (${\sim}7.8$M parameters).
Training uses $\mathrm{Re}{=}1{,}000$ data, Adam optimiser \citep{kingma2015adam}, learning rate $10^{-3}$, batch size 64, weight decay $10^{-4}$, for 10 epochs.
A 50-epoch saturation check confirms that OOD performance plateaus by epoch 10 (seed 42: $33.86\%$ at both 10 and 50 epochs, while IID improves from $3.32\%$ to $1.47\%$), suggesting that OOD-relevant features are learned early.
Autoregressive rollout: predict frame 10, slide window, repeat 10 times.

\subsection{Baseline Fairness}\label{sec:fairness}

\textbf{FNO.}
FNO \citep{li2021fourier} parameterises the evolution operator in Fourier space.
Each spectral layer applies learned multipliers to truncated Fourier modes:
\begin{equation}\label{eq:fno}
  \omega^{(l+1)}(x) = \sigma\!\Big(\mathcal{F}^{-1}\!\big[R_\theta(k) \cdot \hat{\omega}^{(l)}(k)\big](x) + W^{(l)}\omega^{(l)}(x)\Big),
\end{equation}
where $R_\theta(k)$ are learned spectral weights, $W^{(l)}$ is a pointwise linear transform, and $\sigma$ is a nonlinearity.
This global spectral mixing processes all spatial scales simultaneously but does not explicitly preserve local spatial structure.
We use 4 Fourier layers, 12 modes, width 20 (${\sim}466$K parameters), trained for up to 300 epochs with early stopping (patience 20).

\textbf{Protocol.}
All learned predictors are trained on $\mathrm{Re}{\approx}1{,}000$ data only.
Hyperparameters are selected on in-distribution validation without access to $\mathrm{Re}{\approx}10{,}000$ labels.
Training budgets differ across methods (FNO up to 300 epochs, U-Net 10 epochs); we do not claim exhaustive architecture optimisation.
The baselines diagnose cross-regime failure under standard training protocols, not optimal performance per architecture.

\section{Results}\label{sec:results}

\subsection{Main Results}\label{sec:main-results}

Table~\ref{tab:main} presents six methods spanning zero-parameter baselines, retrieval methods, and learned predictors, all evaluated under the same protocol.

\begin{table*}[t]
\caption{Cross-Reynolds prediction at $\mathrm{Re}{=}10{,}000$ (10-step autoregressive, $N{=}500$). $\pm$ denotes standard deviation over 3 seeds.
Bold: best. Underline: second best. All methods trained on $\mathrm{Re}{\approx}1{,}000$ only.}
\label{tab:main}
\vskip 0.1in
\centering
\small
\begin{tabular}{@{}lrcll@{}}
\toprule
Method & Params & OOD (\%) & Type & Representation \\
\midrule
Persistence & 0 & $61.08$ & baseline & -- \\
FNO (3-seed) & 466K & $46.68 \pm 0.11$ & learned predictor & global spectral \\
PCA relay & 0 & $41.77$ & retrieval, no learned dynamics & global linear (36-d) \\
$k$NN-copy & 0 & $41.01$ & retrieval, no learned dynamics & raw field (4{,}096-d) \\
ConvAE-Relay (3-seed) & 1.6M & $\underline{38.34 \pm 0.07}$ & retrieval, source-trained & learned local (64-d) \\
U-Net (3-seed) & 7.8M & $\mathbf{34.72 \pm 0.60}$ & learned predictor & local multi-scale \\
\bottomrule
\end{tabular}
\vskip -0.1in
\end{table*}

\textbf{FNO concentrates at high error.}
FNO reaches $46.68 \pm 0.11\%$ at $\mathrm{Re}{=}10{,}000$ despite strong in-distribution performance.
Three FNO-family variants (vanilla, factorised, scale-consistency loss) all remain in the $46$--$48\%$ range (Appendix~\ref{app:extended}), suggesting a persistent limitation of the tested FNO-family protocols under this shift.

\textbf{Retrieval is competitive without learned dynamics.}
PCA relay ($41.77\%$) and $k$NN-copy ($41.01\%$), both zero-parameter methods, outperform FNO.
Source-regime state matching and dynamics borrowing, without any learned forward map, transfers better than a trained spectral operator.

\textbf{ConvAE-Relay improves with local representation.}
ConvAE-Relay achieves $38.34 \pm 0.07\%$ (independently reproduced within $0.4$\,pp).
The only difference from PCA relay is the matching representation: ConvAE latent versus PCA components.
This provides evidence that the matching representation is a primary driver of the observed improvement under this protocol.

\textbf{U-Net provides learned-side consistent evidence.}
A U-Net predictor achieves $34.72 \pm 0.60\%$, the strongest tested result.
On the canonical FNO benchmark, U-Net underperforms FNO in-distribution in the original comparison \citep{li2021fourier}.
Under cross-Re shift, our U-Net predictor substantially outperforms FNO, suggesting that the architectural features beneficial for in-distribution accuracy (global spectral processing) differ from those for OOD transfer (local multi-scale skip pathways).

\subsection{Representation Ablation}\label{sec:2x2}

Upgrading the matching space from PCA to ConvAE latent improves the best result by $2.37$\,pp ($40.71 \to 38.34$), while the largest update-rule effect within one matching space is $1.06$\,pp.
Raw copy outperforms delta borrowing for PCA ($40.71$ vs $41.77$), while delta outperforms copy for raw fields ($40.21$ vs $41.01$).
ConvAE-latent matching with raw-field delta borrowing is the best overall (Table~\ref{tab:2x2}).
This isolates representation quality as the dominant factor; the update mechanism plays a secondary, representation-dependent role.

\begin{table}[!htb]
\caption{$2{\times}2$ representation ablation (seed 42). Matching quality (rows) dominates update rule (columns).}
\label{tab:2x2}
\vskip 0.1in
\centering
\small
\begin{tabular}{@{}lcc@{}}
\toprule
Matching space & Delta borrow (\%) & Raw copy (\%) \\
\midrule
PCA (36-d) & $41.77$ & $40.71$ \\
Raw field (4{,}096-d) & $40.21$ & $41.01$ \\
\rowcolor{gray!15} ConvAE latent (64-d) & $\mathbf{38.34}$ & -- \\
\bottomrule
\end{tabular}
\vskip -0.1in
\end{table}

\subsection{Bottleneck Decomposition}\label{sec:bottleneck}

Oracle experiments progressively remove bottlenecks to identify where the largest deployable improvements lie (Table~\ref{tab:oracle}).

\begin{table}[!htb]
\caption{Oracle-assisted diagnostics. These are not deployable methods; they are target-informed probes that identify bottleneck contributions.}
\label{tab:oracle}
\vskip 0.1in
\centering
\small
\begin{tabular}{@{}lrl@{}}
\toprule
Variant & OOD (\%) & Removes \\
\midrule
ConvAE-Relay & $38.34$ & -- \\
Oracle matching & $26.49$ & Drift ($\sim$12\,pp) \\
Oracle + $\alpha{=}2.0$ & $24.27$ & + magnitude ($\sim$2\,pp) \\
Oracle magnitude & $23.32$ & + residual mag. \\
\bottomrule
\end{tabular}
\vskip -0.1in
\end{table}

\textbf{Drift is the dominant bottleneck.}
Oracle matching replaces the autoregressive state $\hat{\omega}_t$ with ground truth $\omega_t$ for neighbour retrieval: $j^\star_\mathrm{oracle} = \operatorname*{arg\,min}_{j} d_\phi(\omega_t, \omega^s_j)$.
This yields $26.49\%$, an improvement of ${\sim}12$\,pp.
Drift accounts for approximately 12 percentage points and is the single largest engineering-accessible bottleneck.

\begin{figure}[t]
\centering
\includegraphics[width=\columnwidth]{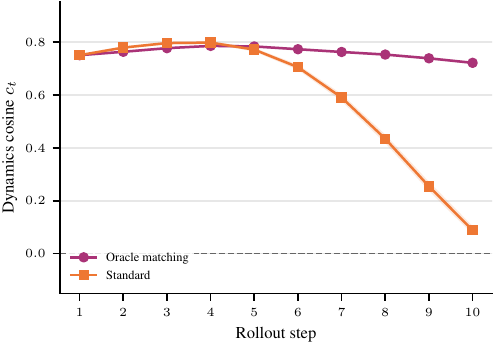}
\caption{Per-step dynamics cosine similarity $c_t = \langle \widehat{\Delta\omega}_t, \Delta\omega_t \rangle / (\lVert\widehat{\Delta\omega}_t\rVert_2 \lVert\Delta\omega_t\rVert_2)$ between borrowed and actual dynamics.
Under oracle matching, similarity remains stable at ${\sim}0.84$; under standard relay, it collapses by step 10.}
\label{fig:cosine}
\end{figure}

\textbf{Dynamics directions are transferable.}
Under oracle matching, the per-step cosine similarity between borrowed source-regime dynamics and actual target-regime dynamics remains stable at ${\sim}0.84$ across all 10 steps.
Under standard relay, this similarity collapses from $0.84$ to $-0.07$ by step 10 (Figure~\ref{fig:cosine}).
The collapse is caused by drift, not by source dynamics being intrinsically wrong: when matching stays on-manifold, borrowed dynamics remain aligned.

\textbf{Magnitude mismatch.}
Scaling borrowed dynamics by a factor $\alpha$, i.e.\ $\hat{\omega}_{t+1} = \omega_t + \alpha \cdot \Delta\omega^s_{j^\star}$, with $\alpha{=}2.0$ reduces error from $26.49\%$ to $24.27\%$; oracle per-step magnitude brings it to $23.32\%$.
Borrowed dynamics magnitudes average $41\%$ of target magnitudes, which is physical: $\mathrm{Re}{=}10{,}000$ fluctuations are stronger.

\textbf{FNO's failure is directional.}
Oracle-optimal scalar rescaling of FNO output yields $46.41\%$, barely reducing the error.
FNO's cross-Re failure is directional, not a trivial magnitude calibration problem.

\subsection{Mechanism Diagnosis by Failed Alternatives}\label{sec:mechanism}

Each ablation tests a hypothesis about how to improve relay, eliminates it, and thereby sharpens the diagnosis.

\textbf{Smart tracking} ($38.21\%$).
\emph{Hypothesis:} Adaptive divergence-based re-matching solves drift.
\emph{Result:} $<0.15$\,pp improvement.
\emph{Elimination:} Re-matching strategy is not the bottleneck.

\textbf{Dynamics-aware AE} ($40.53\%$).
\emph{Hypothesis:} Joint optimisation of matching and dynamics in latent space.
\emph{Result:} Delta geometry improves (norm ratio $0.69 \to 1.02$; direction cosine $0.773 \to 0.888$), but matching error degrades ($15.63\% \to 20.87\%$).
Net result: $2.19$\,pp worse.
\emph{Elimination:} Matching quality and dynamics quality compete within a shared representation.
This is a representation tradeoff: a latent space optimised for dynamics prediction geometry sacrifices nearest-neighbour indexability.

\textbf{$N_\mathrm{ride}$ ablation} ($<0.1$\,pp range).
Ride length has negligible effect.

\textbf{History matching} ($43.32\%$).
\emph{Hypothesis:} Temporal context (10-frame predicted history) helps matching.
\emph{Result:} $5$\,pp worse than baseline.
Under oracle ground-truth history, error drops to $22.83\%$ ($15$\,pp better).
\emph{Elimination:} Phase information is valuable when clean, but autoregressive drift contaminates the history window.

\begin{figure}[t]
\centering
\includegraphics[width=\columnwidth]{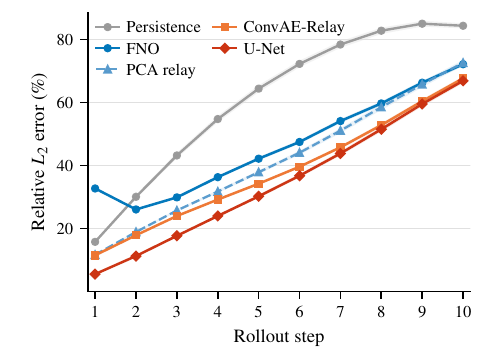}
\caption{Per-step relative $L_2$ error over 10 rollout steps at $\mathrm{Re}{=}10{,}000$.
Shaded bands denote 95\% bootstrap confidence intervals over 500 trajectories (narrow bands indicate low trajectory-level variance).
U-Net maintains the lowest error throughout; all methods show monotonically increasing error consistent with autoregressive drift accumulation.}
\label{fig:perstep}
\end{figure}

\begin{figure}[t]
\centering
\includegraphics[width=\columnwidth]{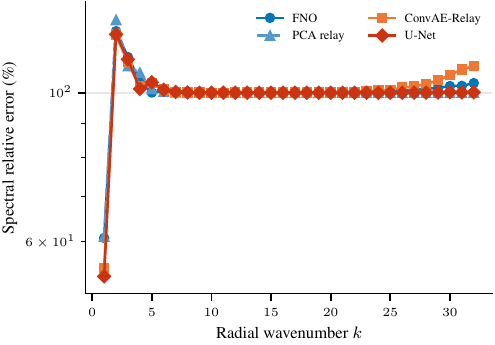}
\caption{Wavenumber-resolved spectral relative error at rollout step 10.
All methods show higher error at low wavenumbers (large-scale dynamics);
U-Net achieves the lowest error across most scales.
Shaded bands denote 95\% bootstrap confidence intervals.}
\label{fig:spectral}
\end{figure}

\subsection{Boundary Characterisation}\label{sec:boundaries}

\textbf{$\mathrm{Re}{=}100{,}000$.}
At $\mathrm{Re}{=}100{,}000$, U-Net remains the strongest tested method ($58.85 \pm 0.32\%$), outperforming FNO ($63.08\%$) by 4\,pp.
However, ConvAE-Relay ($65.65\%$) is worse than FNO, reversing its advantage from the $\mathrm{Re}{=}10{,}000$ regime.
We interpret the relay reversal as a coverage boundary: target states at $\mathrm{Re}{=}100{,}000$ are no longer well indexed by the $\mathrm{Re}{=}1{,}000$ database.
This distinction is important: in the relay-effective regime ($10{\times}$ shift), representation geometry organises which neighbourhoods are useful; outside that regime, support-set coverage dominates and the relay mechanism itself breaks down regardless of representation quality.
U-Net's retained advantage suggests that learned multi-scale representations capture some cross-Re structure even at $100{\times}$ shift, though all methods degrade substantially toward persistence ($64.93\%$).
Appendix Figure~\ref{fig:spectrum} provides a complementary spectral view: the enstrophy spectrum ratio reveals how each method damps or amplifies energy across spatial scales.

\textbf{Long horizon.}
At $T{=}20$, ConvAE-Relay ($66.60\%$) barely outperforms persistence ($68.51\%$), consistent with drift accumulation eroding matching quality.

\textbf{Database size.}
About $75\%$ of improvement from 100 to $4{,}000$ source trajectories is achieved by the first 500, indicating diminishing returns from larger databases under the current representation.

\textbf{U-Net saturation.}
OOD performance saturates within 10 epochs (seed 42: $33.86\%$ at both 10 and 50 epochs; IID: $3.32\% \to 1.47\%$).
Further source-regime training improves in-distribution fit without OOD benefit, suggesting that OOD-relevant features are learned early.

\textbf{U-Net relay.}
U-Net relay variant B ($39.40\%$) is worse than ConvAE-Relay ($38.34\%$) despite U-Net's stronger predictive encoder.
A representation can be predictively sufficient but not metrically indexable: the latent geometry that supports forward prediction need not support nearest-neighbour dynamics retrieval.

\begin{figure}[t]
\centering
\includegraphics[width=\columnwidth]{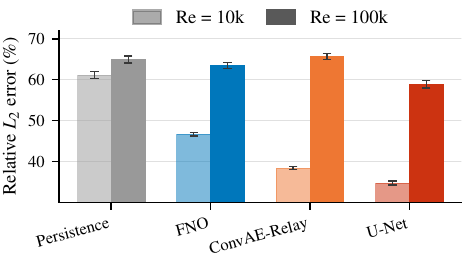}
\caption{Cross-regime comparison at $\mathrm{Re}{=}10{,}000$ and $\mathrm{Re}{=}100{,}000$.
At $10{\times}$ shift, ConvAE-Relay outperforms FNO; at $100{\times}$ shift, this advantage reverses.
U-Net retains its advantage at both regimes.
Error bars denote 95\% bootstrap confidence intervals.}
\label{fig:re100k}
\end{figure}

\section{Discussion and Conclusion}\label{sec:discussion}

The key finding is that cross-Reynolds performance among tested methods is organised by representation geometry, not by whether dynamics are learned or retrieved.
Retrieval methods improve as the matching representation becomes more local and nonlinear (PCA $\to$ $k$NN $\to$ ConvAE).
Learned predictors improve as the architecture provides more local and multi-scale access (FNO $\to$ U-Net).
These two complementary perspectives provide consistent evidence pointing to the same variable.

An independently designed local convolutional predictor without skip connections or residual pathways achieves $48.84\%$, worse than FNO ($46.68\%$) despite using the same local convolutional operations.
This indicates that local spatial processing alone is insufficient; multi-scale information pathways (skip connections) are the critical architectural feature that distinguishes U-Net from less successful local architectures.

\textbf{Empirical lessons for cross-regime solvers.}
By \emph{representation geometry} we mean the neighbourhood structure induced by an encoder $\phi$ and similarity metric: whether nearest neighbours in representation space yield transferable transition vectors under cross-Re shift.
This definition applies directly to retrieval methods (PCA, ConvAE), where the encoder and metric are explicit.
For learned predictors (FNO, U-Net), representation geometry serves as an architectural proxy: the spectral vs.\ local multi-scale distinction describes the inductive bias, not a directly measured latent-space property.
Verifying this proxy through latent-space neighbourhood analysis of U-Net features is a natural next step.
Our results suggest three empirical lessons:
(i) for retrieval-based transfer, locality-preserving nonlinear representations produced more useful nearest neighbours than global linear coordinates on this benchmark;
(ii) locality alone is insufficient; cross-Re prediction benefits from architectures that preserve coarse-to-fine pathways (a local convolutional predictor without skip connections scores $48.84\%$, worse than FNO);
(iii) a representation can be predictively sufficient without being indexable: latent features useful for a forward model need not preserve the neighbourhood structure required for nearest-neighbour transition retrieval.

\textbf{Mechanistic insights.}
U-Net relay underperforming ConvAE relay shows that predictive sufficiency does not imply metric indexability.
Oracle history versus predicted history ($22.83\%$ vs $43.32\%$) shows that phase information is valuable when clean, but destroyed by autoregressive drift.

\textbf{A representation tradeoff: retrieval vs.\ prediction.}
The dynamics-aware AE experiment reveals a non-trivial structural finding: optimising a shared latent space for dynamics prediction geometry (norm ratio $0.69 \to 1.02$; direction cosine $0.773 \to 0.888$) degrades nearest-neighbour matching quality ($15.63\% \to 20.87\%$ reconstruction error), yielding a net $2.19$\,pp increase in OOD error.
A single representation cannot simultaneously optimise for retrieval indexability and forward dynamics prediction.
This tradeoff may generalise beyond the present benchmark: any method using a shared latent space for both retrieval and generation (including video prediction, weather forecasting, and retrieval-augmented language modelling) should face an analogous tension.
We propose this as a testable prediction for future work.

\textbf{What this paper does not claim.}
We do not claim exhaustive architecture optimisation.
We do not claim ConvAE-Relay is parameter-free (the autoencoder has ${\sim}1.6$M parameters trained on source data).
We do not claim that Reynolds-invariant physics has been learned by any tested method.
We do not claim that these findings generalise beyond this benchmark.
We claim that, under a fixed forced 2D Navier--Stokes protocol with $10{\times}$ Re shift, representation geometry is strongly associated with the observed cross-Re error structure among tested methods, and propose this as an organising hypothesis for further investigation.

\textbf{Scope and extensions.}
This study intentionally uses one controlled benchmark (forced 2D NS, $64{\times}64$) to support mechanism-level diagnostics; all claims are scoped accordingly.
The representation-geometry organising principle identified here generates testable predictions for other PDEs (Burgers, 3D turbulence): if the principle holds, local multi-scale architectures should consistently outperform global spectral ones under regime shift across domains.
Our coarse-attention screen established that simply augmenting U-Net with attention under canonical controls does not improve cross-Re transfer ($+0.78$\,pp degradation; Appendix~\ref{app:extended}); investigating whether full Transformer neural operators \citep{cao2021choose, hao2023gnot} exhibit different scaling behaviour would further clarify the role of non-local mixing.
The ${\sim}12$\,pp drift gap quantified by our oracle decomposition provides a precise engineering target: anti-drift strategies such as pushforward training \citep{brandstetter2022message} or diffusion-based refinement \citep{lippe2024pde} could substantially close this gap.
The $20.49$\,pp gap between oracle history ($22.83\%$) and predicted history ($43.32\%$) is even larger than single-frame drift, suggesting that anti-drift mechanisms targeting the temporal context window may yield larger gains than improving the matching representation alone.
Phase-aware temporal encoders and multi-scale relay (per-level delta borrowing in a U-Net feature hierarchy) are promising paths toward representations that remain indexable across wider regime shifts, potentially extending the coverage boundary beyond $\mathrm{Re}{=}10{,}000$.

\section*{Impact Statement}
This paper studies evaluation reliability and regime-shift behaviour for neural PDE solvers. More reliable cross-regime diagnostics may help researchers avoid overstating simulator robustness in scientific and engineering settings. The experiments are limited to an established two-dimensional benchmark and do not introduce a deployable forecasting or control system.


\section*{Acknowledgements}

The author thanks Philip Chan, Jiarui Zhang, Zichen Ma, and
Yongchen Lin for helpful discussions. This work was developed
with assistance from large language models for code generation,
literature review, and drafting; all research design, experimental
decisions, and final interpretations are the author's own.

\bibliography{references}
\bibliographystyle{icml2026}

\newpage
\onecolumn
\appendix

\section{Training Details and Evaluation Audit}\label{app:training}

\textbf{FNO.}
4 Fourier layers, 12 Fourier modes, width 20, ${\sim}466$K parameters.
Adam, lr $10^{-3}$, up to 300 epochs with early stopping (patience 20; best epochs: 31, 25, 52).
IID: $1.11\%$. 3 seeds: $46.68 \pm 0.11\%$.

\textbf{ConvAE.}
4 stride-2 Conv2d layers (1$\to$32$\to$64$\to$128$\to$256), BN, ReLU, linear to 64-d.
Adam, lr $10^{-3}$, 50 epochs.
Reconstruction error ${\sim}1\%$.
Relay: $N_\mathrm{ride}{=}3$, cosine distance, $156{,}000$ database entries.
3 seeds: $38.34 \pm 0.07\%$.

\textbf{U-Net.}
Encoder-decoder with skip connections, circular padding.
Base channels 32$\to$64$\to$128$\to$256$\to$512, ${\sim}7.8$M parameters.
Adam, lr $10^{-3}$, batch 64, weight decay $10^{-4}$, 10 epochs.
IID: $3.00 \pm 0.67\%$. OOD: $34.72 \pm 0.60\%$.

\textbf{Protocol audit.}
All methods evaluated on the same protocol: $\mathrm{Re}{=}10{,}000$, trajectories $0$--$499$, target frames $10$--$19$, mean relative $L_2$.
All reported numbers are post-audit, using a unified frame convention, target window, and averaging method established after the correction of an evaluation indexing error discovered during the project.

\begin{table}[h]
\caption{Protocol audit summary.}
\centering
\small
\begin{tabular}{@{}llcccc@{}}
\toprule
Method & Train Re & Eval Re & Traj. & Frames & Seeds \\
\midrule
FNO & 1k & 10k & 0--499 & 10--19 & 3 \\
PCA relay & 1k & 10k & 0--499 & 10--19 & N/A \\
$k$NN-copy & 1k & 10k & 0--499 & 10--19 & N/A \\
ConvAE-Relay & 1k & 10k & 0--499 & 10--19 & 3 \\
U-Net & 1k & 10k & 0--499 & 10--19 & 3 \\
\bottomrule
\end{tabular}
\end{table}

\section{Extended Results}\label{app:extended}

\textbf{FNO-family band.}
F-FNO (3-seed): $47.49 \pm 0.82\%$.
FNO + scale-consistency loss: $46.65\%$.
All tested FNO-family protocols cluster at $46$--$48\%$.

\textbf{Database size ablation} (seed 42).
100 trajectories: $39.79\%$; 250: $39.16\%$; 500: $38.68\%$; $1{,}000$: $38.60\%$; $2{,}000$: $38.36\%$; $4{,}000$: $38.30\%$.

\textbf{Long-horizon} (seed 42).
$T{=}10$: relay $38.30\%$, PCA $41.77\%$, persistence $61.08\%$.
$T{=}15$: relay $53.76\%$, PCA $56.87\%$, persistence $64.51\%$.
$T{=}20$: relay $66.60\%$, PCA $68.04\%$, persistence $68.51\%$.

\textbf{U-Net saturation} (seed 42).
10 epochs: OOD $33.86\%$, IID $3.32\%$.
50 epochs: OOD $33.86\%$, IID $1.47\%$.

\textbf{U-Net relay 3-variant screen} (seed 42).
Variant A (train with skips, no skips at relay): failed semantic alignment.
Variant B (no skips): $39.40\%$ [95\% CI: 39.00, 39.77].
Variant C (source skips at relay): failed semantic alignment.

\textbf{Local convolutional predictor} (negative result, 2-seed).
$48.84 \pm 0.20\%$ OOD, $1.32 \pm 0.29\%$ IID. No skip connections, no residual pathways. Worse than FNO despite being local, confirming that multi-scale skip pathways, not local convolutions alone, drive U-Net's advantage.

\textbf{$\mathrm{Re}{=}100{,}000$ comparison.}
U-Net (3-seed): $58.85 \pm 0.32\%$.
FNO: $63.08\%$.
Persistence: $64.93\%$.
ConvAE-Relay: $65.65\%$.
U-Net best by $4$\,pp over FNO; relay reverses to worse than FNO.

\textbf{Post-hoc attention screen} (seed 42, canonical controls: batch 64, weight decay $10^{-4}$).
Baseline: $33.86\%$ [CI: 33.36, 34.38].
MHA-only (1B\_attn): $34.64\%$ [CI: 34.15, 35.12], $+0.78$\,pp worse.
At $\mathrm{Re}{=}100{,}000$: baseline $58.99\%$, MHA-only $59.16\%$ ($+0.17$\,pp worse).
Under canonical training controls, coarse-attention augmentation degrades OOD performance, consistent with the finding that better source-regime fitting does not improve cross-Re transfer.

\textbf{Training-pipeline decomposition} (approximate).
FNO: $46.68\%$.
U-Net (batch 32, no weight decay): $37.12\%$, suggesting ${\sim}10$\,pp architectural advantage.
U-Net (batch 64, weight decay $10^{-4}$): $33.86\%$, suggesting ${\sim}3$\,pp additional pipeline effect.
Because batch size, weight decay, and optimiser step count all change together, we treat this as a pipeline effect, not a controlled ablation.

\begin{figure}[h]
\centering
\includegraphics[width=\textwidth]{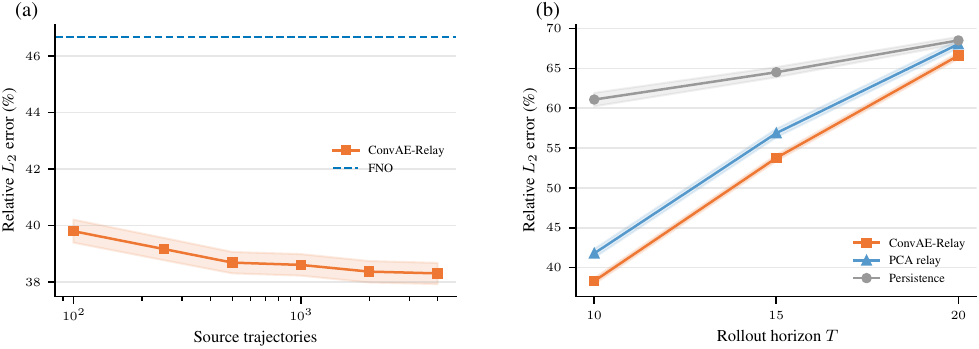}
\caption{Relay scaling behaviour.
\textbf{(a)}~Database size ablation: approximately $75\%$ of improvement is achieved by the first 500 source trajectories, with diminishing returns beyond.
\textbf{(b)}~Long-horizon degradation: by $T{=}20$, ConvAE-Relay converges toward persistence, consistent with drift accumulation eroding matching quality over extended rollouts.
Error bars denote 95\% bootstrap confidence intervals.}
\label{fig:scaling}
\end{figure}

\begin{figure}[h]
\centering
\includegraphics[width=0.5\textwidth]{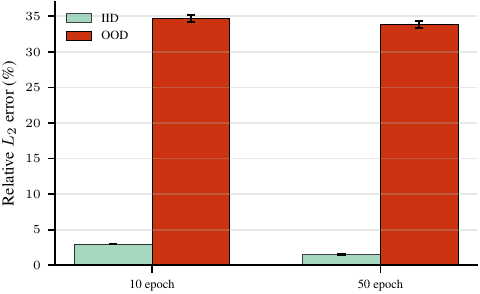}
\caption{U-Net epoch saturation (seed 42). OOD performance at $\mathrm{Re}{=}10{,}000$ saturates by 10 epochs, while in-distribution error continues improving to 50 epochs.
Further source-regime training does not reduce OOD error.}
\label{fig:saturation}
\end{figure}

\begin{figure}[h]
\centering
\includegraphics[width=0.5\textwidth]{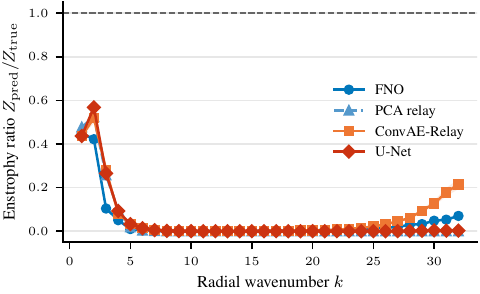}
\caption{Enstrophy spectrum ratio $Z_{\mathrm{pred}}(k)/Z_{\mathrm{true}}(k)$ at rollout step 10.
Values above 1 indicate spectral energy amplification; below 1 indicates damping.
The dashed line marks perfect reconstruction ($\text{ratio}{=}1$).}
\label{fig:spectrum}
\end{figure}

\end{document}